\documentclass[runningheads]{llncs}

\usepackage{eccv}

\usepackage{eccvabbrv}

\usepackage{graphicx}
\usepackage{booktabs}
\usepackage[free-standing-units=true]{siunitx}

\usepackage[accsupp]{axessibility}  

\usepackage{hyperref}

\usepackage{orcidlink}

\usepackage{physics}
\usepackage{tabularx}
\usepackage[normalem]{ulem}
\usepackage{colortbl}
\usepackage[super]{nth}
\usepackage{wrapfig}
\usepackage{footnote}

\let\llncssubparagraph\subparagraph
\let\subparagraph\paragraph
\usepackage[compact]{titlesec}
\let\subparagraph\llncssubparagraph
\titlespacing{\section}{0pt}{2ex}{1ex}
\titlespacing{\subsection}{0pt}{1ex}{0.2ex}
\titlespacing{\subsubsection}{0pt}{0.3ex}{0.1ex}

\hypersetup{
  hidelinks
}

\newcommand{\bbR}{\mathbb{R}}

\newcommand{\fref}[1]{Fig.~\ref{#1}}
\newcommand{\sref}[1]{Sec.~\ref{#1}}
\newcommand{\tref}[1]{Table~\ref{#1}}

\DeclareMathOperator*{\argmin}{argmin}

\definecolor{cvpink}{RGB}{219, 48, 122}
\begin{document}

\title{iMatching: Imperative Correspondence Learning}

\author{
\hspace{-14pt}
Zitong Zhan\inst{1}$^\star$\orcidlink{0009-0003-4111-766X}
\and
Dasong Gao\inst{2}$^\star$\orcidlink{0000-0002-1391-0869} \and
Yun-Jou Lin\inst{3} \and
Youjie Xia\inst{3} \and
Chen Wang\inst{1}\orcidlink{0000-0002-4630-0805}
\hspace{-14pt}
}

\authorrunning{Z.~Zhan et al.}

\institute{SAIR Lab, IAD, CSE, University at Buffalo, Buffalo, NY 14260, USA\\
\email{\{zitongz, chenw\}@sairlab.org} \and
Massachusetts Institute of Technology, Cambridge, MA 02139, USA\\
\email{dasongg@mit.edu}\\
\and
InnoPeak Technology, Palo Alto, CA 94303, USA\\
\email{\{rose.lin, youjie.xia\}@oppo.com}
}

\maketitle

\begin{abstract}
Learning feature correspondence is a foundational task in computer vision, holding immense importance for downstream applications such as visual odometry and 3D reconstruction.
Despite recent progress in data-driven models, feature correspondence learning is still limited by the lack of accurate per-pixel correspondence labels.
To overcome this difficulty, we introduce a new self-supervised scheme, imperative learning (IL), for training feature correspondence.
It enables correspondence learning on arbitrary uninterrupted videos without any camera pose or depth labels, heralding a new era for self-supervised correspondence learning.
Specifically, we formulated the problem of correspondence learning as a bilevel optimization, which takes the reprojection error from bundle adjustment as a supervisory signal for the model. It leads to a mutual improvement between the matching model and the bundle adjustment.
To avoid large memory and computation overhead, we leverage the stationary point to efficiently back-propagate the implicit gradients through bundle adjustment.
Through extensive experiments, we demonstrate superior performance on tasks including feature matching and pose estimation, in which we obtained an average of 30\% accuracy gain over the state-of-the-art matching models. Source code is available at \href{https://github.com/sair-lab/iMatching}{\color{cvpink}github.com/sair-lab/iMatching}.
    \keywords{Matching \and Imperative Learning \and Bilevel Optimization}
\end{abstract}
\renewcommand*{\thefootnote}{\fnsymbol{footnote}}
\footnotetext[1]{Equal contribution.}
\renewcommand*{\thefootnote}{\arabic{footnote}}

\begin{figure*}[t]
    \centering
    \includegraphics[width=\linewidth]{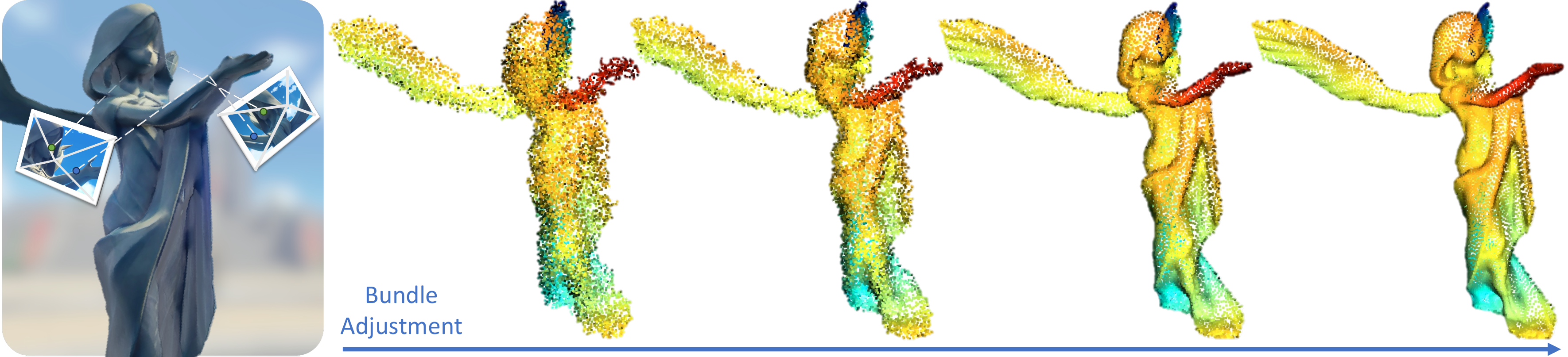}
    \caption{iMatching learns feature correspondence from a bundle adjustment (right), eliminating the need for any ground truth label. The bundle adjustment finds the most geometrically consistent landmark position and camera pose and can be used as a self-supervision signal for feature correspondence networks. }
    \label{fig:teaser_archon}
    \vspace{-13pt}
\end{figure*}

\section{Introduction} \label{sec:intro}

Feature correspondence has long been recognized as a pivotal task in computer vision, acting as a cornerstone for many downstream applications such as visual odometry \cite{xu2023airvo}, 3D reconstruction \cite{stathopoulou2019open}, and object identification \cite{keetha2021airobject}. The essence of accurately matching features across diverse images and viewpoints is critical in constructing coherent and precise representations of the visual world. 

While recent years have witnessed significant advancements in data-driven models, the task remains intrinsically challenging \cite{sarlin2020superglue,chen2022aspanformer,sun2021loftr}, predominantly due to the absence of sufficient pixel-level annotated data for point matching. 
Unlike the labels for other computer vision tasks such as object detection where ground truth annotations can be drawn by hand, annotating feature correspondence for every pixel is almost impossible.
This constraint has inadvertently limited the potential and generalization ability of the state-of-the-art data-driven models \cite{schonberger2017comparative, luo2018geodesc}.
Moreover, the domain gap in visual appearances between training and testing data also greatly discounted model performance \cite{li2020self}. 

To alleviate the data problem, there are three common ways. 
The first way is to simulate correspondence annotation, either from homography transformation \cite{detone2018superpoint} or synthetic environments \cite{wang2020tartanair}. 
However, this method often falls short in capturing real-world variations like changes in lighting and occlusions, leading to the sim-to-real gap. 
The second way is to collect ground truth such as camera pose or depth. 
However, this uses specialized equipment like high-end motion capture systems, usually only available indoors and infeasible for outdoor scenes or unique settings like underground environments. 
Moreover, sensors such as depth cameras and LiDAR tend to offer inaccurate and incomplete depth measurements \cite{lopes2022survey}, which are inadequate for pixel-level matching \cite{schonberger2017comparative}. Alternatively, structure-from-motion (SfM) pipelines could alleviate the usage of physical sensors \cite{li2018megadepth} but still require capturing the scene exhaustively and a long time for preprocessing. 
A third way to address the challenge
has been weakly- and self-supervised methodologies.
SGP \cite{yang2021self} proposes a student-teacher framework for training the CAPS correspondence model \cite{wang2020caps} in a self-supervised manner. 
In SGP, the teacher (RANSAC) and the student model are updated alternately, allowing the teacher to produce pseudo-pose labels for the student model.
Nonetheless, this design is applicable to a limited set of models like CAPS, which only learn from epipolar consistency instead of matching point locations.
Moreover, the teacher and student model training is isolated, leading to potential error propagation, \ie, mistakes made by the teacher model can be propagated to the student model, potentially amplifying the issues.

To fully address the data problem, we explore the third category and propose a new self-supervised end-to-end learning framework, iMatching (\fref{fig:teaser_archon}), for feature correspondence using an emerging technique, imperative learning \cite{yang2023iplanner, fu2024islam}.
For the first time, we formulate the problem of feature matching as a bilevel optimization \cite{ji2021bilevel} where we direct the learning of the model parameters with an optimization procedure, \ie, bundle adjustment (BA), which is itself an optimization process.
Specifically, at the lower level optimization, we update the camera pose and the 3D landmark locations that best explain the model-predicted correspondences, whereas, at the higher level optimization, we update the network parameters such that the correspondence prediction results in more consistent poses and landmark locations.
This new formulation for feature correspondence based on bilevel optimization brings distinctive benefits: The low-level optimization eliminates the need for ground truth geometric labels, thanks to the self-correction mechanism brought by bundle adjustment.
Such a design is flexible, lightweight, and generalizable to any up-to-date matching model.
In the experiments, we show that our method is plug-and-play and significantly boosts the performance of state-of-the-art (SOTA) models \cite{edstedt2023dkm,chen2022aspanformer} by an average of 30\% and a maximum 82\% on pixel-level feature matching on unseen image-only datasets.

One of the challenges of our formulation is model training which requires differentiating the network through low-level optimization (BA).
A popular solution is to backpropagate the errors through the unrolled iterative optimization \cite{jatavallabhula2020slam,teed2021droid}, which is extremely resource-demanding as the temporary variables generated during each iteration need to be stored and traversed for gradient calculation.
Instead, we specially designed our loss function so that we can bypass the iterative process during backward pass while still obtaining the gradient.
This allowed us to employ a black-box, non-differentiable bundle adjustment to ensure the robustness of training while learning end-to-end \textit{as if} it were differentiable.
This end-to-end learnable design also enables the possibility of performing online learning. In summary, our contributions are summarized as follows:
\begin{itemize}
    \item We propose the first imperative learning (IL)-based framework for training feature correspondence models with \textit{uninterrupted videos} but without any ground truth labels such as camera pose or depth.

    \item We design an efficient method for propagating the gradient through the bundle adjustment so that we can solve the optimization without differentiating individual steps of the optimization, making end-to-end training possible.

    \item Through extensive experiments and evaluation on a variety of tasks, we show the superiority of the proposed framework by boosting the performance of the SOTA feature-matching model by 30\% precision on the pose estimation task and 13.6\% on the feature matching task. 
\end{itemize}

\section{Related Work}

Data-driven methods for feature matching categorized as supervised, weakly- or self-supervised will be introduced in \sref{sec:supervised} and \ref{sec:unsupervised}, respectively. 
Additionally, we summarize imperative learning and data-driven models involving bundle adjustment in \sref{sec:il} and \ref{sec:ba}, respectively.

\subsection{Supervised Methods} \label{sec:supervised}

Supervised methods rely on known per-pixel correspondences obtained either from (1) synthetic data or (2) indirect ground truth such as depth images and camera poses. 
The first category uses photo-realistic renderers \cite{wang2020tartanair} or data augmentation techniques to generate training images.
For example, LIFT \cite{yi2016lift} learns from image samples capturing the same view but with strong illumination change. 
SuperPoint \cite{detone2018superpoint} relies on generic image datasets with homographic transformations. 
However, those synthetic data techniques may not generalize well to the real-world scenario. For example, homographic transformations can only model scenes where all the points are coplanar. 

Most of the existing methods \cite{sarlin2020superglue, wang2022lifelong, dusmanu2019d2, revaud2019r2d2, sun2021loftr, chen2022aspanformer,  edstedt2023dkm} belong to the second category, which leverage collected ground truth depth scan and camera pose data from physical sensors, e.g. MoCap, depth camera, LiDAR, or from SfM reconstruction.
However, MoCap is confined to dedicated working zones. Scenes like underground urban sewer systems or mines \cite{oivio} are unable to accommodate complex MoCap devices; depth cameras are limited to indoor environments \cite{lopes2022survey} and provide limited precision, and LiDAR sensors can only provide sparse points. 
As a result, 3D scene data have a significantly smaller scale compared with generic 2D image due to the complexity of data collection. 
3D real world datasets choices are limited. There are only a few datasets such as MegaDepth \cite{li2018megadepth} and ScanNet \cite{dai2017scannet} consist of COLMAP\cite{schoenberger2016sfm} reconstruction from a large number of collected images and have been used for learning feature correspondence, but those datasets are either limited to small-scale environments or lack data diversity \cite{schonberger2017comparative}. In contrast, our method does not explicitly reconstruct a dense 3D model from the complete video sequence and thus is more lightweight and flexible to use.

\subsection{Weakly- and Self-supervised Methods} \label{sec:unsupervised}

Several efforts have focused on weakly-supervised methodologies to address the challenges of obtaining large and diverse training sources. 
For instance, matching labels could be generated from RANSAC-estimated image homography transformation \cite{shen2019ransacflow} or consistency relation between warped images \cite{warpC}. Representation learning on videos \cite{crw, bian2022learning, mckee2022transferrepresentationsvideolabel} is another line of research often evaluated with label propagation tasks but included for completeness. 

We next discuss methods that rely on assumptions in 3D geometry. 
CAPS \cite{wang2020caps} and Patch2Pix \cite{zhou2021patch2pix} eliminate the need for depth information and depend solely on relative camera poses. They leverage the differentiability of matchers so that feature correspondences are corrected by penalizing epipolar distance.

To further remove the dependency on ground truth camera pose labels, self-supervised methods \cite{yang2021self, opflow-bilevel} attempt to jointly estimate the pixel matching and camera poses. For example, SGP\cite{yang2021self} trains CAPS \cite{wang2020caps} using the teacher-student framework.
It introduces RANSAC pose estimation as a teacher to generate pseudo labels for training the student CAPS model.
However, the student and teacher are isolated, which may cause error propagation.

More recently, OmniMotion \cite{omnimotion} proposed a self-supervised test-time optimization that can be applied to individual video sequences and produce per-pixel tracking. The method finds pixel correspondence by building an implicit 3D scene representation and a mapping between image pixels and 3D locations, and such representation is learned by photo rendering loss and point motion smoothness. However, its optimization is difficult in complex scenes, and inference is costly during test time, making it impractical for online usage. In the experiments, we found it requires about 10 hours to test on a 160-frame video. 

Different from previous works, our proposed self-supervised learning framework formulated as bilevel optimization is end-to-end and workable for all models that are directly supervised by pixel locations. This self-supervised learning framework can be applied to a wide variety of models, without altering their original structure, as demonstrated in the experiments.

\subsection{Imperative Learning} \label{sec:il}

Imperative learning (IL) is an emerging self-supervised learning framework based on bilevel optimization. 
It has been used to tackle problems where both data-driven models and geometry problems are involved.
For example, iPlanner \cite{yang2023iplanner} adopted a B-spline interpolation \cite{zhan2023pypose} as the low-level optimization to guide the network to generate smoothed trajectories for path planning; whereas iSLAM \cite{fu2024islam} utilized the pose graph optimization to achieve the reciprocal learning for the front-end and back-end in a simultaneous location and mapping (SLAM) system.
To the best of our knowledge, IL has not been applied to the problem of feature correspondence learning, and iMatching is the first method for self-supervised feature matching using bundle adjustment.

\subsection{Bundle Adjustment in Deep Learning} \label{sec:ba}

Bundle adjustment (BA) is an optimization technique used in computer vision and photogrammetry to simultaneously estimate the 3D coordinates of landmarks and camera parameters by minimizing the reprojection errors between observed and predicted image points.
Recently, several works have tried to integrate BA as part of a neural network.
For example, DROID-SLAM \cite{teed2021droid} uses BA to correct single-view depth estimation during inference so that it is consistent with the geometry shape. 
For training dense depth and camera poses, BA can be used to encourage geometric consistency between pose and depth prediction \cite{shi2019self}. 
However, to make the BA process differentiable, existing methods \cite{teed2021droid,shi2019self,jatavallabhula2020slam} unroll the entire iterative second-order optimization process as part of the auto-differentiation graph.
This incurs a significant computation and memory overhead since the computation graphs of each iteration need to be kept in memory and traversed during backpropagation. 
In contrast, we leverage the stationary point of BA, resulting in an extremely efficient one-step implicit gradient computation. 
As an additional benefit, such a formulation allows us to freely leverage non-differentiable techniques such as alternating pruning and refinement (\sref{sec:formulation:ba}) to enhance the robustness of BA.

\section{Imperative Correspondence Learning}
\label{sec:imperative-correspondence}

We present an overview of iMatching training scheme and its formulation. We then present the matching network the and design of BA.

\begin{figure}[t]
\centering
\begin{minipage}{0.42\textwidth}
\centering
\begin{subfigure}{\textwidth}
\includegraphics[width=\textwidth]{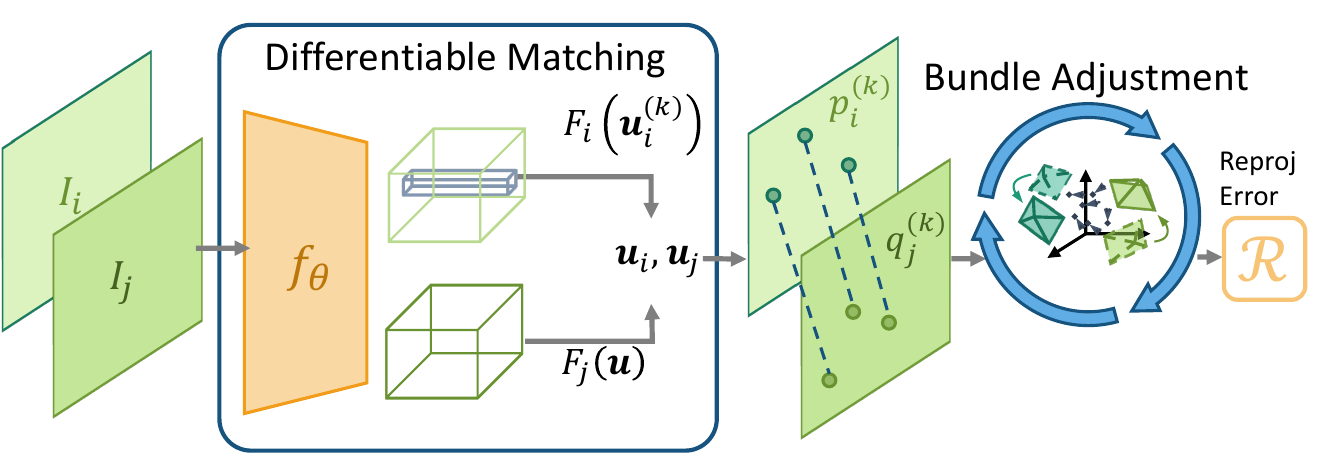}
\caption{System Workflow}
\label{fig:arch}
\end{subfigure}
\begin{subfigure}{\textwidth}
\includegraphics[width=\textwidth]{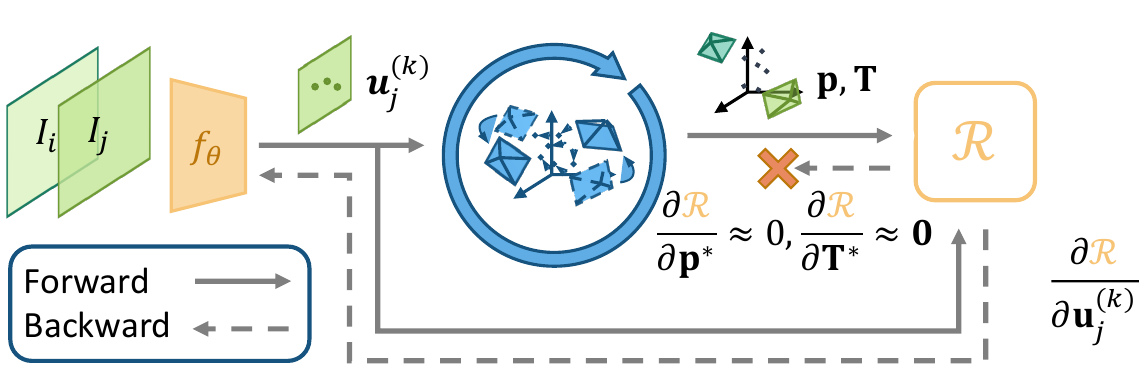}
\caption{Gradient Flow}
\label{fig:bp}
\end{subfigure}
\end{minipage}
\begin{minipage}{0.53\textwidth}
\centering
\begin{subfigure}{\textwidth}
\includegraphics[width=\textwidth]{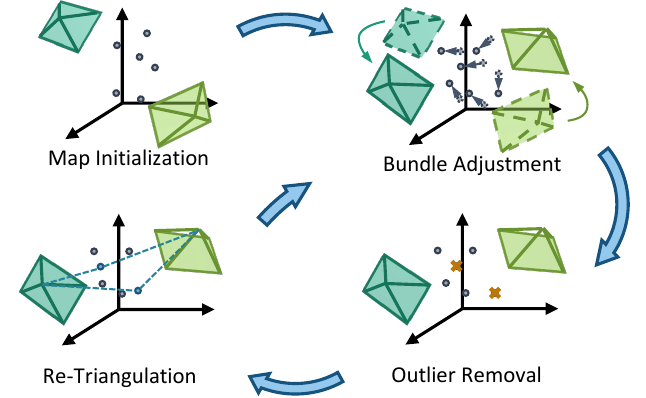}
\caption{Iterative Bundle Adjustment}
\label{fig:ba}
\end{subfigure}
\end{minipage}
\vspace{3pt}
\caption{\textbf{(a)} The iMatching framework for feature correspondence learning consists of a feature correspondence network and a bundle adjustment process, which results in a bilevel optimization. \textbf{(b) Gradient flow in iMatching.} During training, the gradient from the reprojection and epipolar losses is only backpropagated through the 2D feature correspondences $\mathbf{u}_j^{(k)}$ to train the feature matching network. According to our analysis, backpropagation through the robust iterative estimation could be avoided due to the stationarity of optimization outputs $\mathbf{{p}}$ and $\mathbf{T}$.  \textbf{(c) Iterative bundle adjustment} first initializes the camera poses and landmark positions, and then alternates between optimizing reprojection error and outlier rejection until convergence.}
\label{fig:my-figure}
\vspace{-14pt}
\end{figure}

\subsection{Problem Formulation}
\label{sec:formulation}

The architecture of our iMatching training scheme is shown in \fref{fig:arch}, which can be divided into two parts, a feature correspondence network $f_{\boldsymbol{\theta}}$ parameterized by $\boldsymbol{\theta}$ and a BA optimization process represented by \eqref{eqn:lower}. The system is formulated as a bilevel optimization problem, expressed as:
\begin{subequations}\label{eq:bilevel}
\begin{align}
    \min_{\boldsymbol{\theta}}& \;\; \mathcal{R}(f_{\boldsymbol{\theta}}; \mathbf{T}^*, \mathbf{p}^*), \label{eqn:upper} \\ 
    \operatorname{s.t.}& \;\; \mathbf{T}^*, \mathbf{p}^* = \argmin_{{\mathbf{T}, \mathbf{p}}} \mathcal{R}(\mathbf{T}, \mathbf{p}; f_{\boldsymbol{\theta}}),
    \label{eqn:lower}
\end{align}
\end{subequations}
where $\mathcal{R}$ is the reprojection error function; each $\mathbf{T}_i \in \mathbb{SE}(3)$ is a camera pose in Lie group representation; and $\mathbf{p}_l \in \bbR^{3}$ is a 3D landmark position.
Intuitively, at the lower level, BA finds \emph{the most likely poses and landmark positions} given feature correspondence inferred using network parameters $\boldsymbol{\theta}$; whereas at the upper level, it optimizes for \emph{the best parameters} $\boldsymbol{\theta}$ for least reprojection error $\mathcal{R}$ given the estimated poses and landmark positions from the lower level.
This training scheme is named \textit{imperative learning} due to this passive reciprocal optimization between the two levels.
Since both levels only involve minimizing the projection error, it eliminates the requirements in any form of ground truth. 

Each training iteration involves a forward pass of the feature correspondence network and the BA, with their formulations given below, then followed by the optimization of network parameters detailed in \sref{sec:opt}. 

\subsubsection{Feature Correspondence. \vspace{0pt}}
\label{sec:formulation:feat_matching}
Generally, given image pairs $(I_i, I_j)$, the feature correspondence network $f_{\boldsymbol{\theta}}$ produces a set of $K$ pixel correspondences:
\begin{equation}
     f_{\boldsymbol{\theta}}(I_i, I_j) := \{(\textbf{u}_i^{(k)}, \textbf{u}_j^{(k)})\}_{k = 1}^K,
     \label{eq:feat_corr_def}
\end{equation}
where $\textbf{u}_i^{(k)}$ and $\textbf{u}_j^{(k)}$ are the 2D coordinates on image $I_i$ and $I_j$, respectively.
In practice, given a batch of $B$ images, we match each image with its past $n$~$(1 < n < B)$ frames to encourage a stronger geometric constraint.

\subsubsection{Bundle Adjustment. \vspace{0pt}}
\label{sec:formulation:ba}
Given a set of correspondences, the BA process seeks to find the optimal camera poses $\mathbf{T^*} = \{\mathbf{T}_i\}_{i = 1}^B$ and the 3D positions of landmarks $\mathbf{p^*} = \{\mathbf{p}_l\}_{l = 1}^N$ that minimizes the total reprojection error $\mathcal{R}$ defined as:
\begin{align}
    \setlength{\abovedisplayskip}{2pt}
    \setlength{\belowdisplayskip}{2pt}
    \label{eqn:reprojection-error}
    &\mathcal{R}(\mathbf{T}, \mathbf{p}; f_{\boldsymbol{\theta}}) = \sum_{j=1}^{B} \sum_{i = j-n}^{j-1} \sum_{l = 1}^P \norm{\mathbf{u}_i^{(k)} - \pi \left(\mathbf{p}_l; \mathbf{T}_i \right)}_2^2,
\end{align}
with $\pi$ being the reprojection function and $l$ being the index of the landmark $p_l$, which is triangulated from $\textbf{u}_i^{(k)}$ and $\textbf{u}_j^{(k)}$.
We let $B=4$ and $n=2$, so that the training is lightweight to fit in GPU memory, and $\mathcal{R}$ will accumulate projection errors from $5$ frame pairs in total.

The BA can be roughly divided into initialization and pruning stages. 
We first obtain an accurate initialization of poses $\mathbf{T}$ and landmark locations $\mathbf{p}$, and then run an iterative BA scheme that alternates between optimization and outlier rejection.
We next describe design choices making the BA reliable. As the sole source of supervision, its robustness is fundamental to the success of the training. 

\paragraph{Map Initialization.}
We first find a frame pair with the most correspondences and parallax as the anchor, and then triangulate the positions of the landmarks within the pair to obtain the corresponding 3D locations $p$.
For other frames, we estimate their transformations $\mathbf{T}$ with respect to the anchor frames by solving an EPnP \cite{lepetit2009ep} on given triangulated 3D positions and 2D correspondences. Lastly, with each frame having a known camera pose, we triangulate all unused correspondences to obtain the maximum number of landmarks.

\paragraph{Iterative Bundle Adjustment.}
As illustrated in \fref{fig:ba}, we utilize the BA to refine the $\mathbf{T}$ and $\mathbf{p}$ estimated above.
In practice, single-pass BA is found to be prone to poor convergence due to noise-corrupted keypoint detection and wrong feature correspondences \cite{wu2013towards, schoenberger2016mvs}.
To solve this challenge, we follow a procedure similar to that of COLMAP \cite{schoenberger2016sfm} that alternates joint optimization and outlier rejection.
Specifically, after optimizing the BA objective \eqref{eqn:lower} with the Levenberg-Marquardt (LM) optimizer, each landmark goes through a filter and re-triangulation process: 3D locations with large reprojection errors are pruned from the set of landmarks $\mathbf{p}$ at first, and then the surviving ones are updated by re-triangulated landmark locations with the 2D correspondences and refined camera pose.
This alternative procedure runs for 3 iterations.

\subsection{Optimization}\label{sec:opt}

To train the network $f_{\boldsymbol{\theta}}$ via gradient descent $\boldsymbol{\theta} \leftarrow \boldsymbol{\theta} - \eta \pdv{\mathcal{R}}{\boldsymbol{\theta}}$, we need to evaluate the derivative of the reprojection error \eqref{eqn:upper} w.r.t parameters $\boldsymbol{\theta}$ of the network. According to the chain rule, we calculate the derivative $\pdv{\mathcal{R}}{\boldsymbol{\theta}}$ as
\begin{equation}
    \label{eqn:blo-gd-update}
    \pdv{\mathcal{R}}{\boldsymbol{\theta}} = \left( \pdv{\mathcal{R}}{\mathbf{T}^*} \pdv{\mathbf{T}^*}{f_{\boldsymbol{\theta}}} + \pdv{\mathcal{R}}{\mathbf{{p}}^*} \pdv{\mathbf{{p}}^*}{f_{\boldsymbol{\theta}}} + \pdv{\mathcal{R}}{f_{\boldsymbol{\theta}}}\right) \pdv{f_{\boldsymbol{\theta}}}{\boldsymbol{\theta}}.
\end{equation}
However, Equation \eqref{eqn:blo-gd-update} is well defined only if two conditions are satisfied: \textbf{(1)} the matching network $f_{\boldsymbol{\theta}}$ is differentiable, \ie, $\pdv{f_{\boldsymbol{\theta}}}{\theta}$ exists; and \textbf{(2)} the BA in the low-level optimization \eqref{eqn:lower} is differentiable, \ie, $\pdv{\mathbf{{T}}^*}{f_{\boldsymbol{\theta}}}$ and $\pdv{\mathbf{{x}}^*}{f_{\boldsymbol{\theta}}}$ exist.
However, the challenges of satisfying the two conditions are \textbf{(1)} feature matching is normally represented by matching of discrete pixel locations, while differentiating through the integer pixel locations is unstraightforward; \textbf{(2)} solving BA requires iterative, \nth{2}-order optimizers like LM, which computes large Jacobian matrices and its inversion. This already complex process makes back-propagating through it difficult.
We next describe our solutions to resolve the two challenges. 

\subsubsection{Differentiable Image Correspondences. \vspace{0pt}}
\label{sec:diffable-image-corr}

A feature correspondence model $f_{\boldsymbol{\theta}}$ typically consists of three stages: keypoint detection (optional), feature extraction, and matching prediction \cite{li2020dual}.
Feature descriptors are generally derived from feature maps $F_i$ produced by standard backbone networks, through interpolation at the locations of keypoints. The process of matching prediction involves determining the target feature locations based on extracted descriptors.
To make the predicted correspondences differentiable, current methods typically compute the feature locations by expectation \cite{wang2020caps, zhou2021patch2pix} or regression \cite{edstedt2023dkm, chen2022aspanformer}.

\paragraph{Expectation-based Matching Prediction.}
Given a feature coordinate $\mathbf{u}_i^{(k)}$ from $I_i$ and feature maps $F_i, F_j$, the matching coordinate on image $j$ is an expectation over all possible target feature locations weighted by feature similarity
\begin{equation}
    \label{eqn:diff-matching-similarity}
    \mathbf{u}_j^{(k)} = \frac{1}{H W} \sum_{\mathbf{u}^{(l)} \in I_j} \mathbf{u}^{(l)} \cdot S\left( F_i(\mathbf{u}_i^{(k)}), F_j(\mathbf{u}^{(l)}) \right),
\end{equation}
where $S$ can be cosine similarity as an example.

\paragraph{Regression-based Matching Prediction.}
As a more generalized version of expectation-based prediction,
techniques in this category directly regress the amount of feature shift from the feature maps $F_i$ and $F_j$, using a neural network: 
\begin{equation}
    \label{eqn:diff-matching-regression}
    \mathbf{u}_j^{(k)} = g_{\boldsymbol{\theta}'} \left( F_i, F_j; \mathbf{u}_i^{(k)}
    \right),
\end{equation}
where $g_{\boldsymbol{\theta}'}$ can be a transformer \cite{chen2022aspanformer} or a CNN \cite{edstedt2023dkm}.

It is worth noting that iMatching can be used for both categories.
However, there are two non-differentiable cases. 
First, some keypoint detection networks like SuperPoint \cite{detone2018superpoint} have a ranking process with a discrete nature: feature points $\{ \mathbf{u}_i^{(k)} \}_{k = 1}^K$ are typically picked from the pixel coordinates with top $K$ confidence scores. 
To handle such cases, we can either replace the detection network by a differentiable one or freeze its parameters and thus need not propagate the gradient through it. 
Second, some methods like \cite{detone2018superpoint, sarlin2020superglue} use search-based matching prediction. Searching is non-differentiable and out of the scope of discussion.

\subsubsection{Differentiable Bundle Adjustment. \vspace{0pt}}
\label{sec:blo}

An existing widely-used approach to have a differentiable bundle adjustment is to unroll the optimization loop for a fixed number of steps and treat them as part of the forward pass, as done by \cite{jatavallabhula2020slam,teed2021droid,teed2020raft}.
However, such a strategy requires the retention of each iteration's computation graph and the ability to compute the higher-order gradient, which makes this method inapplicable to large-scale problems.

To avoid explicitly computing $\pdv{\mathbf{{T}}^*}{f_{\boldsymbol{\theta}}}$ and $\pdv{\mathbf{{p}}^*}{f_{\boldsymbol{\theta}}}$ by differentiating through the iterative optimization process, we leverage optimization of BA at convergence, which allows us to efficiently backpropagate gradient through the lower level optimization without compromising its robustness.
Concretely, since the optimization of BA is unconstrained, at its stationary point (either locally or globally), we have $\pdv{\mathcal{R}}{\mathbf{T}^*} \approx \mathbf{0}$, $\pdv{\mathcal{R}}{\mathbf{{p^*}}} \approx \mathbf{0}$. Then the gradient \eqref{eqn:blo-gd-update} becomes
\begin{equation}
    \label{eqn:gd-update}
    \pdv{\mathcal{R}}{\boldsymbol{\theta}} = \pdv{\mathcal{R}}{f_{\boldsymbol{\theta}}} \pdv{f_{\boldsymbol{\theta}}}{\boldsymbol{\theta}}.
\end{equation}
This implies that to obtain the gradient descent update \eqref{eqn:lower}, it is sufficient to evaluate the jacobian of the reprojection error \textit{only once} after bundle adjustment and back-propagate through the correspondence argument, treating $\mathbf{{T}^*}$ and $\mathbf{{p}^*}$ as given.
Then, $f_{\boldsymbol{\theta}}$ can be trained with SGD with the gradient given by \eqref{eqn:gd-update}.
\fref{fig:bp} illustrates such a gradient flow.

It is worth noting that \cite[Table 1, N-loop AID]{ji2021bilevel} proved that, as long as the upper-level optimization has a properly small step size, the bilevel optimization is guaranteed to converge, even if the lower-level optimization has not converged. Although \cite{ji2021bilevel} assumes that the lower-level optimization uses \nth{1}-order optimizers like SGD, we empirically found the bilevel optimization can still converge using \nth{2}-order optimizers with just a few iterations.

\subsubsection{Improving Training Convergence. \vspace{0pt}}
\label{sec:additional-techniques}
To further improve the convergence of training, we retain the coarse-to-fine training scheme which is common in feature correspondence models. These models typically first obtain a rough estimate having the same format as in \eqref{eq:feat_corr_def} from low-resolution image features, and then refine the result using a higher-resolution feature or a cropped image patch to produce fine matching.
While the BA in \sref{sec:formulation:ba} only attempts to minimize reprojection error resulting from the fine estimate for the best accuracy, the rough estimate receives the same gradient update in \eqref{eqn:gd-update}.

\section{Experiments}
\label{sec:exp}

We show the effectiveness of our method by presenting the performance gain after finetuning SOTA correspondence models with iMatching on unseen, \emph{image-only} datasets. We experiment with CAPS \cite{wang2020caps}, Patch2Pix \cite{zhou2021patch2pix}, ASpanFormer \cite{chen2022aspanformer}, and DKM \cite{edstedt2023dkm}, and name the models using our iMatching training scheme as iCAPS, iPatch2Pix, iASpan, and iDKM respectively.
We also compare with other self-supervised methods including SGP \cite{yang2021self} and OmniMotion \cite{omnimotion}.

\subsection{Datasets}

\subsubsection{TartanAir} ~\cite{wang2020tartanair} is a large (3TB) and diverse synthetic SLAM dataset.
Collected in photo-realistic simulation environments, the dataset contains precise ground truth depth and pose labels and covers a wide range of scene types 
as well as challenging conditions such as dynamic lighting and adverse weather. 
We use TartanAir to evaluate pixel-level matching accuracy since it is synthesized and thus free of sensor noises. We exclude scenes where ground truth matching is unobtainable, e.g., those containing surfaces with incorrect texture or fast-moving dynamic objects.
We divide each valid scene into subsets of size 8.5:0.5:1 for training, validation, and testing, respectively. 
Note that we apply our iMatching strategy on training sets only in all of our experiments, although iMatching preserves the flexibility for a test time adaptation usage.

\subsubsection{ETH3D-SLAM} ~\cite{eth3d} provides RGBD sequences captured in the real world.
Its scenes primarily consist of small-scale objects with complex details.  We use the dataset for relative pose estimation as it resembles performance in SLAM applications that require frame-by-frame trajectory computation. 
We exclude the scenes where RGB cameras do not provide information in a completely dark environment or are moved while their view is blocked.

We did not use MegaDepth \cite{li2018megadepth} and ScanNet \cite{dai2017scannet} datasets because they only provide discrete images, while iMatching assumes uninterrupted videos to be available. Moreover, these datasets are not suitable for evaluating feature matching because of a lack of pixel-level accuracy, as their ground truth is generated from structure-from-motion pipelines.

\subsection{Feature Matching}
\label{exp:feat_matching}

\subsubsection{Evaluation Protocol. \vspace{0pt}} 
For each method, we report the mean matching accuracy (MMA), \ie, the percentage of predicted correspondences with reprojection errors less than 1, 2, and 5 pixels.
The ground truth correspondences are computed based on the TartanAir's pose and depth map. 

\begin{table*}[t]
  \small
  \setlength{\tabcolsep}{0.14em}
  \renewcommand{\arraystretch}{1.2}
        \caption{Mean matching accuracy (\%) on the TartanAir dataset under error tolerance of 1, 2, and 5 pixels. Our iMatching training scheme consistently boosts the accuracy of feature matching models, with finetuned results shown in ``iCAPS", ``iPatch2Pix", ``iASpan", and ``iDKM". Top performing method under each error tolerance is in bold. } \label{tab:tartanair-mma}
  \resizebox{\textwidth}{!}{
      \begin{tabular}
        {c | ccc !{\color{lightgray}\vrule} ccc | ccc!{\color{lightgray}\vrule}ccc!{\color{lightgray}\vrule}ccc|ccc!{\color{lightgray}\vrule}ccc|ccc
      }
      \toprule
      Scene & \multicolumn{3}{c}{Indoor} & \multicolumn{3}{c}{Outdoor} & \multicolumn{3}{c}{Natural} & \multicolumn{3}{c}{Artificial} & \multicolumn{3}{c}{Mixed} & \multicolumn{3}{c}{Easy} & \multicolumn{3}{c}{Hard} & \multicolumn{3}{c}{Overall} \\ 
      \midrule
      reproj thres & 1px & 2px & 5px & 1px & 2px & 5px & 1px & 2px & 5px & 1px & 2px & 5px & 1px & 2px & 5px & 1px & 2px & 5px & 1px & 2px & 5px & 1px & 2px & 5px \\
      \midrule
      \multicolumn{25}{l}{{(a) Non-differentiable Matcher}} \\
      \midrule
      SIFT & 22.6 & 31.7 & 41.9 & 22.9 & 31.0 & 38.3 & 16.0 & 23.2 & 30.5 & 24.9 & 33.8 & 42.7 & 24.1 & 32.5 & 39.7 & 23.7 & 32.5 & 40.8 & 21.4 & 29.4 & 37.4 & 22.8 & 31.2 & 39.3 \\
      ORB & 20.8 & 40.2 & 58.5 & 16.5 & 32.0 & 45.9 & 10.2 & 21.2 & 33.0 & 20.1 & 38.4 & 55.0 & 18.5 & 35.9 & 50.9 & 18.8 & 36.5 & 52.1 & 16.7 & 32.4 & 47.2 & 17.7 & 34.4 & 49.6 \\
      SuperGlue & 39.6 & 64.2 & 85.0 & 38.2 & 62.4 & 80.9 & 27.2 & 49.0 & 71.0 & 41.9 & 66.8 & 85.6 & 40.4 & 65.4 & 83.2 & 40.7 & 65.8 & 84.1 & 36.7 & 60.2 & 80.2 & 38.6 & 62.9 & 82.1 \\
      R2D2 & 30.6 & 48.6 & 64.8 & 31.4 & 48.6 & 61.0 & 25.2 & 41.4 & 55.1 & 32.9 & 50.5 & 64.7 & 32.0 & 50.1 & 62.4 & 32.3 & 50.6 & 64.3 & 29.1 & 45.7 & 59.6 & 31.1 & 48.6 & 62.2 \\
      \midrule
      \multicolumn{25}{l}{{(b) \textbf{Pretrained Supervised} Models of End-to-end Differentiable Matcher }} \\
      \midrule
      CAPS & 26.6 & 62.3 & 87.3 & 27.1 & 63.3 & 87.3 & 21.1 & 52.7 & 81.0 & 28.0 & 64.6 & 88.1 & 29.0 & 67.1 & 90.0 & 28.4 & 66.2 & 89.4 & 25.5 & 60.1 & 85.3 & 26.9 & 63.0 & 87.3 \\
      Patch2Pix & 0.7 & 3.5 & 23.6 & 0.9 & 4.6 & 25.8 & 1.0 & 4.9 & 26.1 & 0.8 & 4.1 & 24.9 & 0.8 & 4.3 & 25.0 & 0.9 & 4.6 & 26.6 & 0.8 & 4.0 & 23.9 & 0.8 & 4.3 & 25.2 \\
      ASpanFormer & 51.5 & 76.9 & 94.2 & 51.5 & 76.7 & 93.3 & 43.7 & 70.1 & 90.6 & 52.2 & 77.8 & 94.3 & 55.4 & 79.3 & 94.2 & 55.8 & 80.9 & 95.2 & 47.4 & 72.9 & 92.0 & 51.5 & 76.8 & 93.6 \\
      DKM & 60.3 & 76.4 & 91.9 & 65.2 & 80.1 & 93.4 & 58.4 & 75.2 & 91.1 & 64.0 & 79.0 & 93.0 & 67.1 & 81.7 & 94.3 & 67.8 & 82.7 & 94.9 & 59.0 & 74.8 & 91.0 & 63.8 & 79.0 & 93.0 \\
      \midrule
      \multicolumn{25}{l}{{(c) Self-supervised Baseline}} \\
      \midrule
      SGP & 44.8 & 70.5 & 88.7 & 47.7 & 71.8 & 88.7 & 36.6 & 60.7 & 83.1 & 47.5 & 72.6 & 89.2 & 52.5 & 76.5 & 91.6 & 50.8 & 75.5 & 90.8 & 43.1 & 67.6 & 86.7 & 46.8 & 71.4 & 88.7 \\
      OmniMotion &  38.4 & 62.6 & 86.3 & 28.2 & 52.0 & 78.8 & 26.6 & 49.7 & 78.8 & 33.1 & 57.0 & 81.1 & 29.6 & 54.1 & 81.0 & 33.8 & 58.5 & 83.1 & 28.3 & 51.5 & 78.6 & 30.7 & 54.6 & 80.6 \\
      \midrule
      \multicolumn{25}{l}{\textbf{(d) Models Using Our iMatching Training Scheme (Self-supervised)}} \\
      \midrule
      \textbf{iCAPS (Ours)} & 47.5 & 70.9 & 88.3 & 49.7 & 72.9 & 89.7 & 40.4 & 63.8 & 85.6 & 50.1 & 73.3 & 89.4 & 53.0 & 76.3 & 91.5 & 53.0 & 76.0 & 91.0 & 45.3 & 68.9 & 87.7 & 49.0 & 72.3 & 89.3 \\
      \textbf{iPatch2Pix (Ours)} & 6.1 & 17.3 & 42.1 & 7.5 & 20.6 & 44.5 & 7.3 & 20.2 & 43.8 & 5.8 & 17.4 & 42.5 & 9.3 & 23.4 & 46.3 & 8.3 & 22.9 & 49.0 & 6.0 & 16.7 & 39.0 & 7.1 & 19.7 & 43.8 \\
      \textbf{iASpan (Ours)} & {60.1} & \textbf{79.0} & \textbf{95.0} & {57.8} & {78.7} & {94.2} & {51.7} & {73.6} & {91.8} & {58.6} & {78.9} & \textbf{95.1} & {62.7} & {82.0} & \textbf{94.9} & {63.1} & {82.9} & \textbf{96.0} & {54.0} & {74.9} & \textbf{92.9} & {58.4} & {78.8} & \textbf{94.4} \\
      \textbf{iDKM (Ours)} & \textbf{62.9} & {78.9} & {93.1} & \textbf{66.6} & \textbf{81.4} & \textbf{94.4} & \textbf{60.7} & \textbf{77.5} & \textbf{92.6} & \textbf{65.9} & \textbf{80.9} & {94.1} & \textbf{68.0} & \textbf{82.6} & \textbf{94.9} & \textbf{69.6} & \textbf{84.5} & {95.8} & \textbf{60.6} & \textbf{76.4} & {92.1} & \textbf{65.5} & \textbf{80.7} & {94.0} \\
      \bottomrule
      \end{tabular}
      }
      \vspace{-15pt}
\end{table*}

\begin{figure*}[t]
    \centering
    \includegraphics[width=\linewidth]{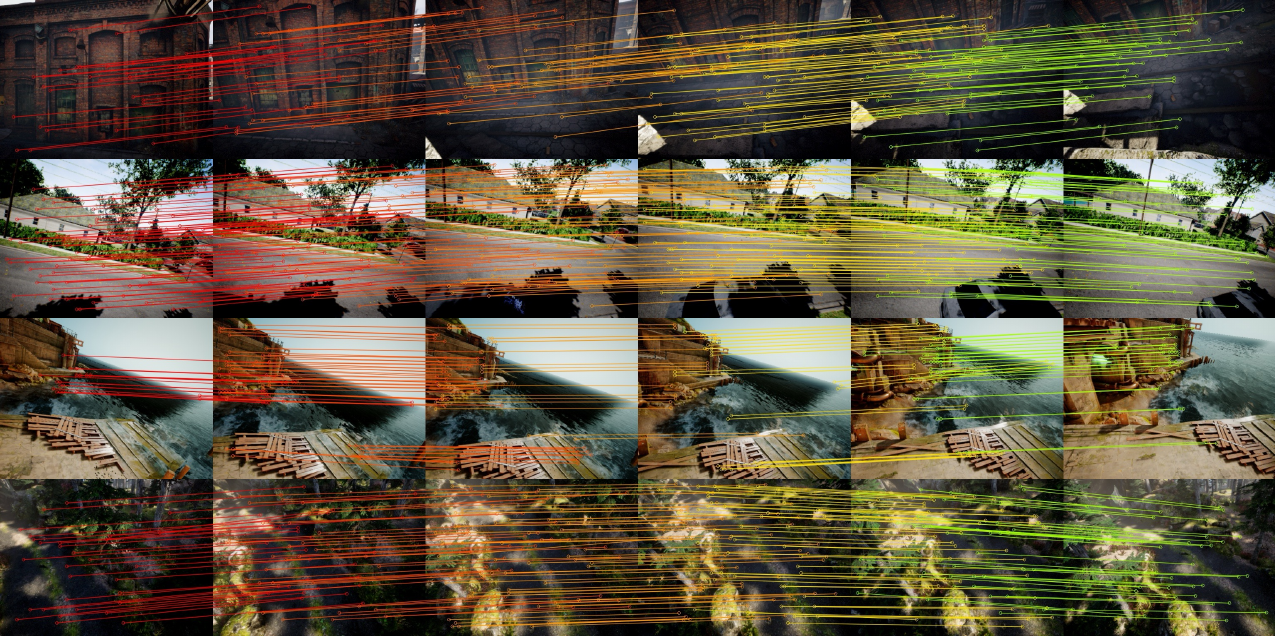}
    \caption{\textbf{iASpan qualitative results}. Each row shows an uninterrupted video with 10\% of the correspondences. Best viewed digitally.}
    \label{fig:wrap_tta}
\vspace{-25pt}
\end{figure*}

\subsubsection{Baselines. \vspace{0pt}} 
We compare against most of the widely used and state-of-the-art models. 
We divide them into three categories: (a) non-differentiable models including SIFT \cite{sift}, ORB \cite{orb}, SuperGlue \cite{sarlin2020superglue}, and R2D2 \cite{revaud2019r2d2}, (b) the state-of-the-art models including CAPS \cite{wang2020caps}, Patch2Pix \cite{zhou2021patch2pix}, ASpanFormer \cite{chen2022aspanformer}, and DKM \cite{edstedt2023dkm}, and (c) self-supervised methods, SGP \cite{yang2021self} and OmniMotion \cite{omnimotion}.
SGP is specifically designed for CAPS \cite{wang2020caps}.
For fairness, both SGP and our iMatching are initialized with the exact same pretrained weights as provided by \cite{wang2020caps}.   
Additionally, we compare with OmniMotion. 
Since OmniMotion can only be applied at test time, we train OmniMotion directly on each testing sequence. 
To ensure OmniMotion isn't negatively impacted by large camera motions in TartanAir's long testing sequence, we trim videos into smaller segments of 160 frames each. 

\subsubsection{Results. \vspace{0pt}}
We report the MMA of selected methods in \tref{tab:tartanair-mma}.
It can be seen that methods trained with iMatching greatly outperform previous handcrafted, supervised, and self-supervised methods. 
Comparing categories (a) and (b), we demonstrate that self-supervision can further enhance data-driven methods, which already outperform handcrafted features in our zero-shot scenario.   

Compared with the pretrained supervised methods in (b), iCAPS in (d) offers 1.82$\times$ of the accuracy of the pretrained CAPS at 1-pixel error tolerance.
\fref{fig:qual} compares the CAPS and iCAPS model on simple point cloud registration tasks, with ground truth depth scan used to highlight pose estimation accuracy. The pretrained model leaves obvious misalignments even if BA is used to promote consensus of camera poses, which indicates flaws in both the model accuracy and robustness of a normal BA, but our bilevel optimization successfully eliminates the artifacts. 
Patch2Pix and ASpanFormer are two models showing weak and strong zero-shot performance, respectively. Our iASpan is capable of improving the already strong ASpanFormer model by 13.6\%, and more qualitative results are available in \fref{fig:wrap_tta}. Our iDKM demonstrates the strongest pixel-level accuracy. Additionally, we observe a strong performance boost of 8.9$\times$ on Patch2Pix, even though it's a poorly performed model initially. This shows that our bilevel optimization formulation is robust against poor initialization.

In addition, our method outperforms SGP by a 5\% higher performance gain, as indicated by (c) and iCAPS in (d). OmniMotion is primarily impacted by aggressive camera motions in hard videos and potentially the highly non-convex nature of optimization as the authors mentioned. Our method has a relatively obvious advantage on Natural scenes. They contain repeating texture patterns and complex lighting conditions, \eg the vegetation in the \nth{4} sample in \fref{fig:wrap_tta}.  These conditions typically cause frequent mismatched correspondence and thus highlight the importance of outlier rejection capability in self-supervised systems. 
We conclude that our experiments on CAPS, Patch2Pix, ASpanFormer, and DKM show the strong adaptability of iMatching by covering and demonstrating performance gain on major types of end-to-end differentiable models.

\subsubsection{Runtime. \vspace{0pt}}
We observed that our method consumes 4.5s for a single training sample, using the CAPS model. The total training time on a single TartanAir scene consisting of 2248 frames is 40 mins. For comparison, SGP takes 110 \minute to complete training, as it requires more iterations to converge. OmniMotion has quadratic computation complexity during test-time adaptation which involves computing pairwise flow of all frames. The same sequence costs more than 2 days on inference. 
In contrast, our method does not alter the architecture of the off-the-shelf models, thus retaining the original model's inference efficiency.

\subsubsection{Additional Baseline. \vspace{0pt}}
We further compare to a fully supervised baseline using DKM, where we perform a COLMAP reconstruction on each sequence and train DKM in a fully supervised manner from the COLMAP-generated label. Details of the COLMAP
experiments on feature matching and the following relative pose estimation task can be found in the supplementary material.

\begin{figure}[t]
\centering
\begin{minipage}{0.33\textwidth}
\centering
\begin{subfigure}{\textwidth}
\includegraphics[width=\textwidth]{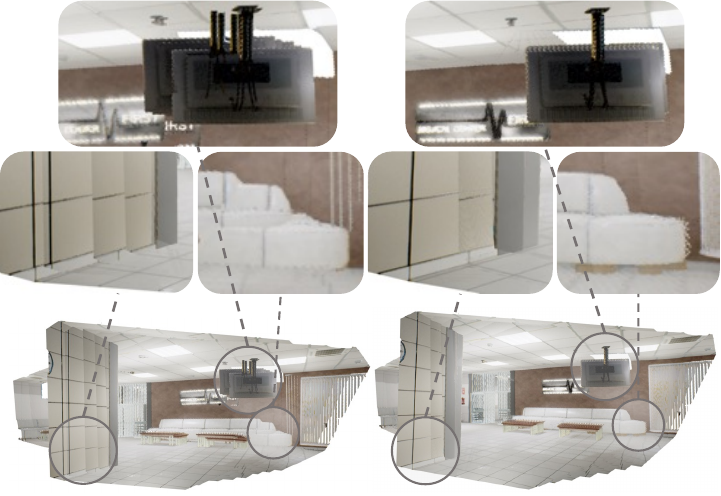}
\caption{3D Reconstruction}
\label{fig:qual}
\end{subfigure}
\end{minipage}
\begin{minipage}{0.60\textwidth}
\centering
\begin{subfigure}{\textwidth}
\includegraphics[width=\textwidth]{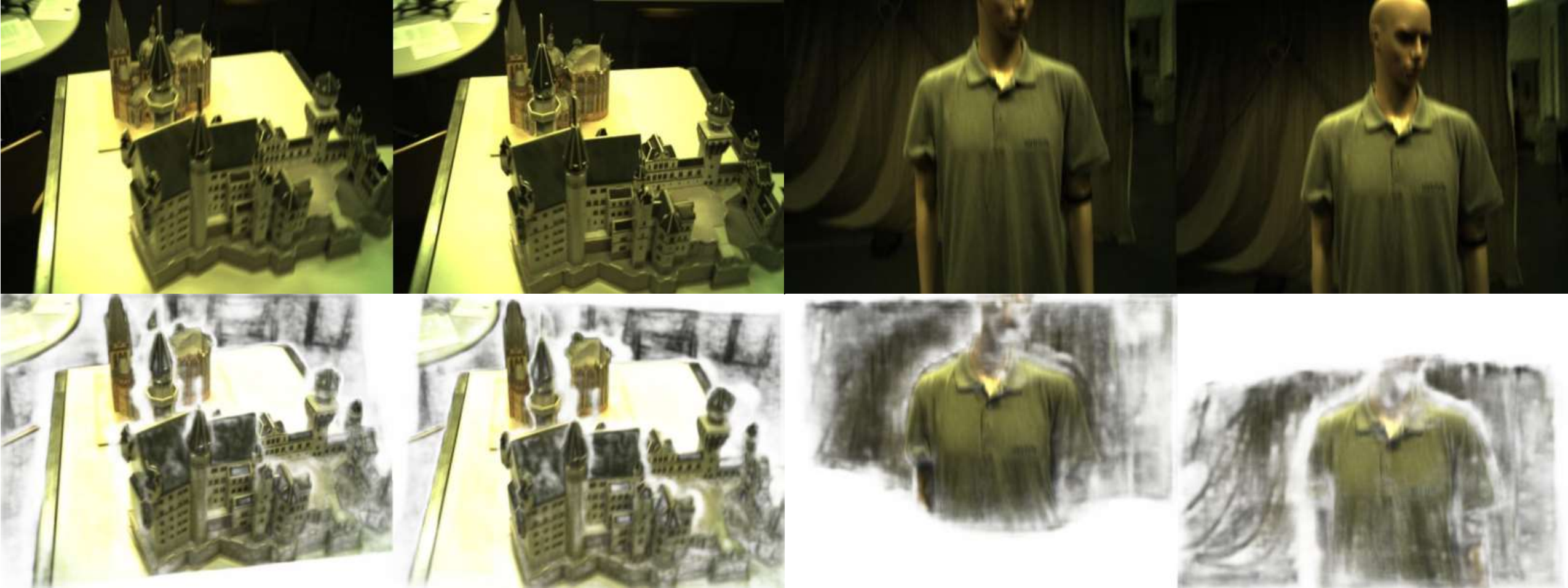}
\caption{iDKM image warping}
\label{fig:wrap_eth}
\end{subfigure}
\end{minipage}
\caption{\textbf{(a) Pose estimation comparison between pretrained and finetuned model.} The bottom point clouds are reconstructed by aligning TartanAir RGBD scans using RANSAC-estimated poses derived from feature correspondence. The pretrained CAPS model (left) creates artifacts on the sofa and monitor due to inaccurately estimated poses and falsely aligned depth map, while such artifacts are not presented in our finetuned model (right). Best viewed digitally. \textbf{(b). iDKM Qualitative Results}. The top row show raw image pairs $I_i$ and $I_j$. These qualitative results contain restored RGB pixel values of one image by transferring pixel color given the estimated feature correspondences from the other image. Our iDKM accurately identifies matching on the features best explained by object shape. Best viewed digitally.}
\vspace{-10pt}
\end{figure}

\subsection{Relative Pose Estimation}
\label{exp:pose}

\begin{table*}[t]
  \centering
  \small
  \setlength{\tabcolsep}{0.26em}
  \caption{\textbf{Pose estimation accuracy on ETH3D-SLAM.} 
  The best results are highlighted in bold. 
  The column ``SGP'' is the accuracy of the finetuned model of CAPS using the self-supervised SGP training scheme.
  The columns ``CAPS'' and ``DKM'' are the accuracies of the pretrained models from their original works.
  The columns ``iCAPS' and ``iDKM'' are accuracies of CAPS and DKM trained by our iMatching self-supervised learning, with a percentage of improvements under 5\textdegree~threshold calculated in ``\%$\uparrow$". }
  \renewcommand{\arraystretch}{1.5}
    \resizebox{\textwidth}{!}{
      \begin{tabular}{c|ccc|ccc!{\color{lightgray}\vrule}ccc!{\color{lightgray}\vrule}cccc|ccc!{\color{lightgray}\vrule}cccc}
      \toprule
      
      Method & \multicolumn{3}{c|}{SuperGlue~\cite{sarlin2020superglue}} & \multicolumn{3}{c}{SGP~\cite{yang2021self}} & \multicolumn{3}{c}{CAPS~\cite{wang2020caps}} & \multicolumn{4}{c|}{\textbf{iCAPS (Ours)}} & \multicolumn{3}{c}{DKM~\cite{edstedt2023dkm}} & \multicolumn{4}{c}{\textbf{iDKM (Ours)}} \\
      \midrule
      AUC & 5\textdegree & 10\textdegree & 20\textdegree  & 5\textdegree & 10\textdegree & 20\textdegree  & 5\textdegree & 10\textdegree & 20\textdegree  & 5\textdegree & 10\textdegree & 20\textdegree & \% $\uparrow$ & 5\textdegree & 10\textdegree & 20\textdegree  & 5\textdegree & 10\textdegree & 20\textdegree & \% $\uparrow$ \\
      \midrule
      cables & 66.9 & 72.6 & 75.4 & 62.0 & 67.9 & 70.9 & 67.4 & 73.9 & 77.1 & \textbf{69.0} & \textbf{76.3} & \textbf{80.0} & 2.4\% & 59.2 & 63.7 & 66.0 & \textbf{84.4} & \textbf{91.6} & \textbf{95.2} & 42.6\% \\
      camera\_shake & 64.5 & 78.0 & 86.1 & 68.4 & 82.1 & 89.8 & 63.0 & 77.8 & 87.0 & \textbf{72.3} & \textbf{84.1} & \textbf{90.8} & 14.8\% & 65.9 & 72.9 & 76.5 & \textbf{67.7} & \textbf{74.2} & \textbf{77.4} & 2.7\% \\
      ceiling & 81.0 & 86.6 & 89.7 & 78.9 & 85.0 & 88.1 & 81.4 & 87.8 & 91.1 & \textbf{83.9} & \textbf{91.0} & \textbf{94.6} & 3.1\% & 59.4 & 62.8 & 64.6 & \textbf{76.6} & \textbf{83.8} & \textbf{87.5} & 29.0\% \\
      desk & 77.1 & 84.5 & 88.4 & 73.5 & 82.3 & 87.0 & 72.8 & 82.2 & 87.3 & \textbf{74.4} & \textbf{83.9} & \textbf{88.9} & 2.2\% & 82.8 & 85.3 & 86.8 & \textbf{82.9} & \textbf{85.5} & \textbf{86.9} & 0.1\% \\
      desk\_changing & 71.3 & 77.2 & 80.4 & 69.7 & 76.8 & 80.4 & 73.5 & 81.6 & 85.8 & \textbf{74.3} & \textbf{83.1} & \textbf{87.8} & 1.1\% & 61.2 & 62.9 & 63.7 & \textbf{75.3} & \textbf{86.1} & \textbf{91.8} & 23.0\% \\
      einstein & 57.1 & 62.0 & 64.8 & 66.7 & 72.4 & 75.8 & 67.8 & 74.1 & 77.8 & \textbf{69.8} & \textbf{76.6} & \textbf{80.5} & 2.9\% & 36.6 & 38.2 & 39.4 & \textbf{63.4} & \textbf{68.4} & \textbf{71.3} & 73.2\% \\
      einstein\_GLC & 42.3 & 46.1 & 48.4 & 51.7 & 57.4 & 60.7 & 51.3 & 56.7 & 60.3 & \textbf{53.6} & \textbf{59.3} & \textbf{63.0} & 4.5\% & 35.5 & 41.0 & 45.9 & \textbf{53.1} & \textbf{62.1} & \textbf{69.9} & 49.6\% \\
      mannequin & 76.2 & 80.6 & 83.0 & 80.1 & 84.9 & 87.5 & 80.2 & 85.3 & 88.0 & \textbf{83.9} & \textbf{89.5} & \textbf{92.5} & 4.6\% & 59.5 & 60.9 & 61.6 & \textbf{81.1} & \textbf{88.1} & \textbf{91.6} & 36.3\% \\
      mannequin\_face & 69.1 & 71.0 & 71.9 & 73.2 & 76.3 & 77.9 & 73.4 & 76.8 & 78.5 & \textbf{77.2} & \textbf{81.3} & \textbf{83.3} & 5.2\% & 53.9 & 54.3 & 54.5 & \textbf{71.3} & \textbf{73.7} & \textbf{74.8} & 32.3\% \\
      planar & 67.7 & 78.9 & 84.5 & 65.6 & 79.5 & 86.7 & 67.1 & 81.6 & 88.9 & \textbf{69.3} & \textbf{83.3} & \textbf{91.0} & 3.3\% & 62.9 & 70.6 & 74.4 & \textbf{71.3} & \textbf{81.1} & \textbf{86.4} & 13.4\% \\
      plant & 78.5 & 82.2 & 84.1 & 74.4 & 80.2 & 83.0 & 71.9 & 78.5 & 81.7 & \textbf{81.0} & \textbf{88.0} & \textbf{91.5} & 12.7\% & 86.5 & 88.9 & 90.0 & \textbf{88.4} & \textbf{91.1} & \textbf{92.4} & 2.2\% \\
      plant\_scene & 72.6 & 77.2 & 79.6 & 71.0 & 76.4 & 79.2 & 71.8 & 77.7 & 80.7 & \textbf{80.2} & \textbf{86.8} & \textbf{90.2} & 11.7\% & 44.2 & 45.2 & 45.7 & \textbf{83.5} & \textbf{90.7} & \textbf{94.3} & 88.9\% \\
      sfm\_lab\_room & 87.7 & 93.8 & 96.9 & \textbf{90.6} & \textbf{95.3} & \textbf{97.6} & 86.4 & 93.3 & 96.6 & {90.4} & {95.2} & \textbf{97.6} & 4.6\% & 92.0 & 94.4 & 95.6 & \textbf{94.2} & \textbf{97.1} & \textbf{98.6} & 2.4\% \\
      sofa & 69.2 & 76.9 & 81.4 & 70.8 & 79.6 & 84.2 & 70.8 & 79.6 & 84.1 & \textbf{72.2} & \textbf{81.5} & \textbf{86.9} & 2.0\% & 52.9 & 54.2 & 54.8 & \textbf{80.1} & \textbf{89.1} & \textbf{94.9} & 51.4\% \\
      table & 65.3 & 68.8 & 70.6 & 65.3 & 69.4 & 71.4 & 70.5 & 75.3 & 77.7 & \textbf{77.4} & \textbf{83.3} & \textbf{86.3} & 9.8\% & 28.6 & 29.4 & 29.8 & \textbf{78.2} & \textbf{90.0} & \textbf{94.5} & 173.4\% \\
      vicon\_light & 69.4 & 76.0 & 79.6 & 72.7 & 81.1 & 85.6 & \textbf{73.2} & \textbf{82.0} & \textbf{86.6} & {71.8} & {81.3} & {86.4} & -1.9\% & 66.5 & 70.1 & 72.1 & \textbf{74.5} & \textbf{80.3} & \textbf{83.3} & 12.0\% \\
      large\_loop & 69.2 & 75.1 & 78.1 & 69.9 & 75.8 & 78.8 & 73.2 & 79.3 & 82.4 & \textbf{78.6} & \textbf{86.2} & \textbf{90.3} & 7.4\% & 42.5 & 45.1 & 46.4 & \textbf{77.7} & \textbf{86.8} & \textbf{91.5} & 82.8\% \\
      \midrule
      \textbf{Overall} & 69.7 & 75.8 & 79.0 & 70.9 & 77.8 & 81.4 & 71.5 & 79.0 & 83.0 & \textbf{75.3} & \textbf{83.0} & \textbf{87.1} & 5.3\% & 58.2 & 61.2 & 62.8 & \textbf{76.7} & \textbf{83.5} & \textbf{87.2} & 31.8\% \\
      \bottomrule
    \end{tabular}
  }
  \label{tab:eth3d}
  \vspace{-20pt}
\end{table*}

We next conduct another task, relative pose estimation, which is a crucial application in computer vision, to demonstrate the adaptability of the iMatching training scheme.
We follow \cite{chen2022aspanformer} to recover the camera pose by solving the essential matrix from the predicted correspondences using RANSAC.

\subsubsection{Baselines. \vspace{0pt}}
To evaluate the performance gain after training, we use both expectation- and regression-based methods with pre-trained weights, as represented by CAPS \cite{wang2020caps} and DKM \cite{edstedt2023dkm}, respectively.
Again, we compare against SGP \cite{yang2021self} by training  CAPS with iMatching.
The non-differentiable SuperGlue model \cite{sarlin2020superglue} is included for reference. We selected these models because they are either detector-based or dense flow models, and thus can provide correspondence at arbitrary pixel locations, which is crucial for applications such as SLAM or VO.

\subsubsection{Accuracy. ~}
A pose estimate is considered to be correct if the angular error in both rotation and translation falls below thresholds of 5$^\circ$, 10$^\circ$, and 20$^\circ$, and we report the area-under-curve (AUC) of the pose error in \tref{tab:eth3d}. iMatching shows superior performance over supervised and self-supervised methods. 
Data-driven methods perform strongly, except for the DKM. 
Despite demonstrating the strongest performance under supervised settings on MegaDepth and ScanNet \cite{edstedt2023dkm}, as well as top pretrained accuracy in our previous feature matching, DKM is behind SuperGlue and CAPS by a large margin on the unseen ETH3D-SLAM scenes. 
However, iMatching can boost its accuracy by 30\% compared with the pretrained, making it the strongest among all methods. 
From image warping results in \fref{fig:wrap_eth}, it shows consistent per-pixel matching quality.

\subsubsection{Comparison with SGP. ~}
Pose estimation accuracy improvement brought by SGP is only observed in 5 scenes and is overall negligible.
This is primarily because the CAPS model has strong generalizability in the task but the simple supervision formulation in CAPS already has saturated performance under the evaluation metrics: SGP training can only encourage overall consensus between correspondence following the epipolar constraint regardless of actual 3D shape, so the effectiveness of SGP is not as obvious as what we observed in feature matching. In contrast, our design allows us to further improve the already strong CAPS model by 5\%. Our iDKM experiment also verifies that iMatching is less likely to be vulnerable to initially worse pose estimation accuracy, as its iterative BA process brings additional robustness to correct error that occurs in map initialization.
Therefore, we conclude that our imperative learning works pragmatically and effectively, thanks to its carefully designed robust BA. 

\subsubsection{Additional evaluation on outdoor scene. ~} To show that our method is generalizable to outdoor environments as well, we finetune the two top-performing methods, ASpanFormer and DKM, on the KITTI360 dataset. We partition the \begin{table}[t]
    \small
    \centering
    \caption{Pose estimation on KITTI360}
    \vspace{-8pt}
    \resizebox{0.5\columnwidth}{!}
    {
    \begin{tabular}{c|ccc|c|ccc}
        \toprule
        Method & $5^\circ$ & $10^\circ$ & $20^\circ$ & Method & $5^\circ$ & $10^\circ$ & $20^\circ$   \\
        \midrule
        ASpanFormer & 74.4 & 86.3 & 93.0  & DKM & 91.8 & 95.9 & 98.0 \\
        \textbf{iASpan (Ours)} & \textbf{80.6} & \textbf{90.1} & \textbf{95.1} & \textbf{iDKM (Ours)} & \textbf{92.6} & \textbf{96.3} & \textbf{98.1}  \\
        \bottomrule
    \end{tabular}
    }
    \label{tab:kitti}
    \vspace{-20pt}
\end{table}
entire dataset using the same ratio as used in TartanAir for training and testing. We evaluate pose AUC using the dataset's ground truth camera pose. Results in \tref{tab:kitti} show that our iMatching brought 8.3\% improvement under 5$^\circ$ error tolerance. The DKM experiment shows less obvious improvement as the model already has saturated performance.

\section{Conclusion}
We explore an exciting new direction in self-supervised feature correspondence learning, \ie imperative learning,  using bundle adjustment as a supervision signal, formulated as a bilevel optimization problem, and learning feature correspondence from uninterrupted video sequence without any form of ground truth labels. 
We highlight the simple and efficient optimization technique used to differentiate through the complex bundle adjustment process. 
We experimented with the state-of-the-art models and demonstrated obvious performance leaps in tasks including feature matching and relative pose estimation, empirically verifying the importance of our robust bundle adjustment. This paper makes an attractive starting point for integrating the learning process of feature correspondence to the underlying optimization in downstream applications such as visual odometry and 3D reconstruction in an online manner. 
\section*{Acknowledgements}
This work was in part supported by the ONR award N00014-24-1-2003. Any opinions, findings, conclusions, or recommendations expressed in this paper are those of the authors and do not necessarily reflect the views of the ONR. The authors wish to express their gratitude for the generous gift funding provided by Cisco Systems Inc. and InnoPeak Technology Inc.
\bibliographystyle{splncs04}
\bibliography{main,publications}

\begin{thebibliography}{10}
\providecommand{\url}[1]{\texttt{#1}}
\providecommand{\urlprefix}{URL }
\providecommand{\doi}[1]{https://doi.org/#1}

\bibitem{bian2022learning}
Bian, Z., Jabri, A., Efros, A.A., Owens, A.: Learning pixel trajectories with multiscale contrastive random walks. In: Proceedings of the IEEE/CVF Conference on Computer Vision and Pattern Recognition (CVPR) (June 2022)

\bibitem{chen2022aspanformer}
Chen, H., Luo, Z., Zhou, L., Tian, Y., Zhen, M., Fang, T., McKinnon, D., Tsin, Y., Quan, L.: Aspanformer: Detector-free image matching with adaptive span transformer. In: Computer Vision--ECCV 2022: 17th European Conference, Tel Aviv, Israel, October 23--27, 2022, Proceedings, Part XXXII. pp. 20--36. Springer (2022)

\bibitem{dai2017scannet}
Dai, A., Chang, A.X., Savva, M., Halber, M., Funkhouser, T., Nie{\ss}ner, M.: Scannet: Richly-annotated 3d reconstructions of indoor scenes. In: Proceedings of the IEEE conference on computer vision and pattern recognition. pp. 5828--5839 (2017)

\bibitem{detone2018superpoint}
DeTone, D., Malisiewicz, T., Rabinovich, A.: Superpoint: Self-supervised interest point detection and description. In: Proceedings of the IEEE conference on computer vision and pattern recognition workshops. pp. 224--236 (2018)

\bibitem{dusmanu2019d2}
Dusmanu, M., Rocco, I., Pajdla, T., Pollefeys, M., Sivic, J., Torii, A., Sattler, T.: D2-net: A trainable cnn for joint detection and description of local features. arXiv preprint arXiv:1905.03561  (2019)

\bibitem{edstedt2023dkm}
Edstedt, J., Athanasiadis, I., Wadenbäck, M., Felsberg, M.: {DKM}: Dense kernelized feature matching for geometry estimation. In: IEEE Conference on Computer Vision and Pattern Recognition (2023)

\bibitem{fu2024islam}
Fu, T., Su, S., Lu, Y., Wang, C.: {iSLAM}: Imperative {SLAM}. IEEE Robotics and Automation Letters (RA-L)  (2024), \url{https://arxiv.org/pdf/2306.07894.pdf}

\bibitem{crw}
Jabri, A., Owens, A., Efros, A.A.: Space-time correspondence as a contrastive random walk. Advances in Neural Information Processing Systems  (2020)

\bibitem{jatavallabhula2020slam}
Jatavallabhula, K.M., Iyer, G., Paull, L.: {$\triangledown$}slam: Dense slam meets automatic differentiation. In: 2020 IEEE International Conference on Robotics and Automation (ICRA). pp. 2130--2137. IEEE (2020)

\bibitem{ji2021bilevel}
Ji, K., Yang, J., Liang, Y.: Bilevel optimization: Convergence analysis and enhanced design. In: International conference on machine learning. pp. 4882--4892. PMLR (2021)

\bibitem{opflow-bilevel}
Jiang, S., Campbell, D., Liu, M., Gould, S., Hartley, R.: Joint unsupervised learning of optical flow and egomotion with bi-level optimization. In: 2020 International Conference on 3D Vision (3DV). pp. 682--691 (2020). \doi{10.1109/3DV50981.2020.00078}

\bibitem{oivio}
Kasper, M., McGuire, S., Heckman, C.: {A Benchmark for Visual-Inertial Odometry Systems Employing Onboard Illumination}. In: Intelligent Robots and Systems (IROS) (2019)

\bibitem{keetha2021airobject}
Keetha, N.V., Wang, C., Qiu, Y., Xu, K., Scherer, S.: {AirObject}: A temporally evolving graph embedding for object identification. In: IEEE/CVF Conference on Computer Vision and Pattern Recognition (CVPR) (2022), \url{https://arxiv.org/pdf/2111.15150}

\bibitem{lepetit2009ep}
Lepetit, V., Moreno-Noguer, F., Fua, P.: Ep n p: An accurate o (n) solution to the p n p problem. International journal of computer vision  \textbf{81},  155--166 (2009)

\bibitem{li2020self}
Li, S., Wang, X., Cao, Y., Xue, F., Yan, Z., Zha, H.: Self-supervised deep visual odometry with online adaptation. In: Proceedings of the IEEE/CVF Conference on Computer Vision and Pattern Recognition. pp. 6339--6348 (2020)

\bibitem{li2020dual}
Li, X., Han, K., Li, S., Prisacariu, V.: Dual-resolution correspondence networks. Advances in Neural Information Processing Systems  \textbf{33},  17346--17357 (2020)

\bibitem{li2018megadepth}
Li, Z., Snavely, N.: Megadepth: Learning single-view depth prediction from internet photos. In: Proceedings of the IEEE conference on computer vision and pattern recognition. pp. 2041--2050 (2018)

\bibitem{lopes2022survey}
Lopes, A., Souza, R., Pedrini, H.: A survey on rgb-d datasets. arXiv preprint arXiv:2201.05761  (2022)

\bibitem{sift}
Lowe, D.G.: Distinctive image features from scale-invariant keypoints. Int. J. Comput. Vision  \textbf{60}(2),  91--110 (Nov 2004). \doi{10.1023/B:VISI.0000029664.99615.94}, \url{http://dx.doi.org/10.1023/B:VISI.0000029664.99615.94}

\bibitem{luo2018geodesc}
Luo, Z., Shen, T., Zhou, L., Zhu, S., Zhang, R., Yao, Y., Fang, T., Quan, L.: Geodesc: Learning local descriptors by integrating geometry constraints. In: Proceedings of the European conference on computer vision (ECCV). pp. 168--183 (2018)

\bibitem{mckee2022transferrepresentationsvideolabel}
McKee, D., Zhan, Z., Shuai, B., Modolo, D., Tighe, J., Lazebnik, S.: Transfer of representations to video label propagation: Implementation factors matter (2022), \url{https://arxiv.org/abs/2203.05553}

\bibitem{revaud2019r2d2}
Revaud, J., Weinzaepfel, P., De~Souza, C., Pion, N., Csurka, G., Cabon, Y., Humenberger, M.: R2d2: repeatable and reliable detector and descriptor. arXiv preprint arXiv:1906.06195  (2019)

\bibitem{orb}
Rublee, E., Rabaud, V., Konolige, K., Bradski, G.R.: Orb: An efficient alternative to sift or surf. 2011 International Conference on Computer Vision pp. 2564--2571 (2011), \url{https://api.semanticscholar.org/CorpusID:206769866}

\bibitem{sarlin2020superglue}
Sarlin, P.E., DeTone, D., Malisiewicz, T., Rabinovich, A.: Superglue: Learning feature matching with graph neural networks. In: Proceedings of the IEEE/CVF conference on computer vision and pattern recognition. pp. 4938--4947 (2020)

\bibitem{schonberger2017comparative}
Schonberger, J.L., Hardmeier, H., Sattler, T., Pollefeys, M.: Comparative evaluation of hand-crafted and learned local features. In: Proceedings of the IEEE conference on computer vision and pattern recognition. pp. 1482--1491 (2017)

\bibitem{schoenberger2016sfm}
Sch\"{o}nberger, J.L., Frahm, J.M.: Structure-from-motion revisited. In: Conference on Computer Vision and Pattern Recognition (CVPR) (2016)

\bibitem{schoenberger2016mvs}
Sch\"{o}nberger, J.L., Zheng, E., Pollefeys, M., Frahm, J.M.: Pixelwise view selection for unstructured multi-view stereo. In: European Conference on Computer Vision (ECCV) (2016)

\bibitem{eth3d}
Sch\"ops, T., Sattler, T., Pollefeys, M.: {BAD SLAM}: Bundle adjusted direct {RGB-D SLAM}. In: Conference on Computer Vision and Pattern Recognition (CVPR) (2019)

\bibitem{shen2019ransacflow}
Shen, X., Darmon, F., Efros, A.A., Aubry, M.: Ransac-flow: generic two-stage image alignment. In: 16th European Conference on Computer Vision (2020)

\bibitem{shi2019self}
Shi, Y., Zhu, J., Fang, Y., Lien, K., Gu, J.: Self-supervised learning of depth and ego-motion with differentiable bundle adjustment. arXiv preprint arXiv:1909.13163  (2019)

\bibitem{stathopoulou2019open}
Stathopoulou, E.K., Welponer, M., Remondino, F.: Open-source image-based 3d reconstruction pipelines: Review, comparison and evaluation. The International Archives of the Photogrammetry, Remote Sensing and Spatial Information Sciences, Volume XLII-2/W17 pp. 331--338 (2019)

\bibitem{sun2021loftr}
Sun, J., Shen, Z., Wang, Y., Bao, H., Zhou, X.: Loftr: Detector-free local feature matching with transformers. In: Proceedings of the IEEE/CVF conference on computer vision and pattern recognition. pp. 8922--8931 (2021)

\bibitem{teed2020raft}
Teed, Z., Deng, J.: Raft: Recurrent all-pairs field transforms for optical flow. In: European conference on computer vision. pp. 402--419. Springer (2020)

\bibitem{teed2021droid}
Teed, Z., Deng, J.: Droid-slam: Deep visual slam for monocular, stereo, and rgb-d cameras. Advances in Neural Information Processing Systems  \textbf{34},  16558--16569 (2021)

\bibitem{warpC}
Truong, P., Danelljan, M., Yu, F., Gool, L.V.: Warp consistency for unsupervised learning of dense correspondences. In: 2021 {IEEE/CVF} International Conference on Computer Vision, {ICCV} 2021, Montreal, QC, Canada, October 10-17, 2021. pp. 10326--10336. {IEEE} (2021). \doi{10.1109/ICCV48922.2021.01018}, \url{https://doi.org/10.1109/ICCV48922.2021.01018}

\bibitem{wang2022lifelong}
Wang, C., Qiu, Y., Gao, D., Scherer, S.: Lifelong graph learning. In: IEEE/CVF Conference on Computer Vision and Pattern Recognition (CVPR) (2022), \url{https://arxiv.org/pdf/2009.00647}

\bibitem{omnimotion}
Wang, Q., Chang, Y.Y., Cai, R., Li, Z., Hariharan, B., Holynski, A., Snavely, N.: Tracking everything everywhere all at once. ICCV  (2023)

\bibitem{wang2020caps}
Wang, Q., Zhou, X., Hariharan, B., Snavely, N.: Learning feature descriptors using camera pose supervision. In: European Conference on Computer Vision. pp. 757--774. Springer (2020)

\bibitem{wang2020tartanair}
Wang, W., Zhu, D., Wang, X., Hu, Y., Qiu, Y., Wang, C., Hu, Y.H., Kapoor, A., Scherer, S.: Tartanair: A dataset to push the limits of visual slam. In: IEEE/RSJ International Conference on Intelligent Robots and Systems (IROS). pp. 4909--4916 (2020), \url{https://arxiv.org/pdf/2003.14338}

\bibitem{wu2013towards}
Wu, C.: Towards linear-time incremental structure from motion. In: 2013 International Conference on 3D Vision-3DV 2013. pp. 127--134. IEEE (2013)

\bibitem{xu2023airvo}
Xu, K., Hao, Y., Yuan, S., Wang, C., Xie, L.: {AirVO}: An illumination-robust point-line visual odometry. In: IEEE/RSJ International Conference on Intelligent Robots and Systems (IROS) (2023), \url{https://arxiv.org/pdf/2212.07595.pdf}

\bibitem{yang2023iplanner}
Yang, F., Wang, C., Cadena, C., Hutter, M.: {iPlanner}: Imperative path planning. In: Robotics: Science and Systems (RSS) (2023), \url{https://arxiv.org/pdf/2302.11434.pdf}

\bibitem{yang2021self}
Yang, H., Dong, W., Carlone, L., Koltun, V.: Self-supervised geometric perception. In: Proceedings of the IEEE/CVF Conference on Computer Vision and Pattern Recognition. pp. 14350--14361 (2021)

\bibitem{yi2016lift}
Yi, K.M., Trulls, E., Lepetit, V., Fua, P.: Lift: Learned invariant feature transform. In: European conference on computer vision. pp. 467--483. Springer (2016)

\bibitem{zhan2023pypose}
Zhan, Z., Li, X., Li, Q., He, H., Pandey, A., Xiao, H., Xu, Y., Chen, X., Xu, K., Cao, K., Zhao, Z., Wang, Z., Xu, H., Fang, Z., Chen, Y., Wang, W., Fang, X., Du, Y., Wu, T., Lin, X., Qiu, Y., Yang, F., Shi, J., Su, S., Lu, Y., Fu, T., Dantu, K., Wu, J., Xie, L., Hutter, M., Carlone, L., Scherer, S., Huang, D., Hu, Y., Geng, J., Wang, C.: {PyPose} v0.6: The imperative programming interface for robotics. In: IEEE/RSJ International Conference on Intelligent Robots and Systems (IROS) Workshop (2023), \url{https://arxiv.org/abs/2309.13035}

\bibitem{zhou2021patch2pix}
Zhou, Q., Sattler, T., Leal-Taixe, L.: Patch2pix: Epipolar-guided pixel-level correspondences. In: Proceedings of the IEEE/CVF conference on computer vision and pattern recognition. pp. 4669--4678 (2021)

\end{thebibliography}

\title{
Supplementary Material for \\
iMatching: Imperative Correspondence Learning} 

\author{
\hspace{-14pt}
Zitong Zhan\inst{1}$^\star$\orcidlink{0009-0003-4111-766X}
\and
Dasong Gao\inst{2}$^\star$\orcidlink{0000-0002-1391-0869} \and
Yun-Jou Lin\inst{3} \and
Youjie Xia\inst{3} \and
Chen Wang\inst{1}\orcidlink{0000-0002-4630-0805}
\hspace{-14pt}
}

\authorrunning{Z.~Zhan et al.}

\institute{SAIR Lab, IAD, CSE, University at Buffalo, Buffalo, NY 14260, USA\\
\email{\{zitongz, chenw\}@sairlab.org} \and
Massachusetts Institute of Technology, Cambridge, MA 02139, USA\\
\email{dasongg@mit.edu}\\
\and
InnoPeak Technology, Palo Alto, CA 94303, USA\\
\email{\{rose.lin, youjie.xia\}@oppo.com}
}

\maketitle

\section{TartanAir Scene Category Definition}

The video sequences in TartanAir are categorized into groups based on their characteristics. The exact definition is included below:

\vspace{15pt}
\resizebox{0.925\columnwidth}{!}
{\begin{tabular}{|l|l|}
\hline
\textbf{Categories} & \textbf{Scenes} \\
\hline
Indoors & \texttt{carwelding}, \texttt{hospital}, \texttt{japanesealley}, \texttt{office}, \texttt{office2} \\
Outdoors & \texttt{abandonedfactory}, \texttt{abandonedfactory\_night}, \texttt{amusement}, \\
& \texttt{endofworld}, \texttt{gascola}, \texttt{ocean}, \texttt{neighborhood}, \texttt{oldtown}, \\ 
 & \texttt{seasonsforest\_winter}, \texttt{seasidetown}, \texttt{soulcity}, \texttt{westerndesert} \\
Natural & \texttt{gascola}, \texttt{ocean}, \texttt{seasonsforest\_winter} \\
Artificial & \texttt{abandonedfactory}, \texttt{abandonedfactory\_night}, \texttt{carwelding},  \\ & \texttt{endofworld}, \texttt{hospital}, \texttt{japanesealley}, \texttt{office}, \texttt{office2}, \texttt{soulcity} \\
Mixed & \texttt{amusement}, \texttt{neighborhood}, \texttt{oldtown}, \texttt{seasidetown}, \texttt{westerndesert} \\
\hline
\end{tabular}
}
\vspace{10pt}

\section{Ablation Study}

In \tref{tab:abla},  we perform ablation studies to verify the effectiveness of our design decisions in bundle adjustment. The CAPS pretrained model is included for reference. 
First, we compared our design of BA with that of COLMAP. The experiment is to replace our implementation of camera pose estimation, triangulation, and bundle adjustment with the COLMAP's reconstruction. The COLMAPtriangulates landmarks from the feature correspondence model's prediction, and the results are used to train the model's correspondence output. We select sequences where COLMAP can succeed. However, it shows weaker performance because COLMAP fails when there isn't sufficient parallax between frames. 

Additionally, we study the required number of iterations when running the BA. The 1-step and 3-steps experiments are to limit the number of LM iterations to 1 and 3 respectively. We also run bundle adjustment without outlier rejection, as documented in column ``w/o rej". The full system is reported as iCAPS. 
We conclude both outlier rejection and sufficient LM iterations are required for the best performance.

\begin{table*}[h]
    \centering
    \caption{Ablation study on bundle adjustment module.}
    \resizebox{0.7\columnwidth}{!}
    {
      \renewcommand{\arraystretch}{1.2}
    \begin{tabular}{c|c|c|cccc|c}
        \toprule
        sequence & tolerance & CAPS & \textbf{1} step & \textbf{3} steps & w/o rej & \textbf{iCAPS (Ours)} & COLMAP \\
        \midrule
        Amusement & 1px & 29.8 & 50.4 & 52.7 & 50.6 & 59.7 & 31.4\\
        Easy & 2px & 73.1 & 84.6 & 85.4 & 84.0 & 87.0 & 73.7\\            \hline
        Hospital & 1px & 26.3 & 44.8 & 45.9 & 46.5 & 50.2 & 25.8\\
        Hard & 2px & 59.9 & 68.0 & 68.9 & 68.8 & 69.6 & 58.8\\        
        \hline
        Oldtown & 1px & 33.5 & 53.5 & 56.6 & 42.6 & 59.2 & 32.5\\
        Easy & 2px & 74.2 & 80.1 & 80.2 & 73.7 & 80.7 & 69.4\\        
        \hline
        SeasonForest & 1px & 17.1 & 21.9 & 22.3 & 21.9 & 29.4 & 17.8\\
        Easy & 2px & 43.7 & 47.9 & 48.3 & 47.6 & 50.7 & 43.9\\        
        \bottomrule
    \end{tabular}
    }
    \label{tab:abla}
\end{table*}

\section{COLMAP Fully Supervised Baseline}

\begin{wraptable}{r}{0.5\linewidth}
  \centering
  \small
  \setlength{\tabcolsep}{0.26em}
  \vspace{-30pt}
  \caption{Comparison with additional baseline on the {pose estimation task} }
  \renewcommand{\arraystretch}{1.5}
    \resizebox{0.5\textwidth}{!}{
      \begin{tabular}{c|ccc!{\color{lightgray}\vrule}ccc}
      \toprule
      
      Method & \multicolumn{3}{c|}{COLMAP} & \multicolumn{3}{c}{iDKM (Ours)} \\
      \midrule
      AUC & 5\textdegree & 10\textdegree & 20\textdegree  & 5\textdegree & 10\textdegree & 20\textdegree  \\
    \midrule
cables & 88.2 & 92.2 & 94.3 & 84.4 & 91.6 & 95.2 \\
camera\_shake & 42.2 & 63.0 & 76.4 & 67.7 & 74.2 & 77.4 \\
ceiling & 84.6 & 91.4 & 94.9 & 76.6 & 83.8 & 87.5 \\
desk & 87.1 & 92.0 & 94.8 & 82.9 & 85.5 & 86.9 \\
desk\_changing & 92.5 & 95.6 & 97.3 & 75.3 & 86.1 & 91.8 \\
einstein & 67.9 & 74.4 & 78.2 & 63.4 & 68.4 & 71.3 \\
einstein\_GLC & 48.8 & 59.6 & 67.0 & 53.1 & 62.1 & 69.9 \\
mannequin & 82.9 & 89.1 & 92.2 & 81.1 & 88.1 & 91.6 \\
mannequin\_face & 92.6 & 95.3 & 96.7 & 71.3 & 73.7 & 74.8 \\
planar & 63.3 & 81.6 & 90.8 & 71.3 & 81.1 & 86.4 \\
plant & 94.4 & 97.2 & 98.6 & 88.4 & 91.1 & 92.4 \\
plant\_scene & 90.2 & 94.4 & 96.4 & 83.5 & 90.7 & 94.3 \\
sfm\_lab\_room & 76.8 & 85.9 & 90.5 & 94.2 & 97.1 & 98.6 \\
sofa & 84.6 & 90.8 & 93.9 & 80.1 & 89.1 & 94.9 \\
table &      &      &      & 78.2 & 90.0 & 94.5 \\
vicon\_light & 77.4 & 86.7 & 91.5 & 74.5 & 80.3 & 83.3 \\
large\_loop & 74.1 & 85.6 & 91.5 & 77.7 & 86.8 & 91.5 \\
\textbf{Overall} & 77.9 & 85.9 & 90.3 & 76.7 & 83.5 & 87.2 \\
      \bottomrule
    \end{tabular}
  }
  \label{tab:supp:eth3d}
  \vspace{-20pt}
\end{wraptable}

Another baseline is to train the correspondence matching model in a supervised manner from the 3D model and camera poses constructed by COLMAP with video images and MVS enabled.
The difference between the baseline and our approach is that our iMatching does not explicitly reconstruct a dense 3D model from the complete video sequence. Instead, iMatching only triangulates sparse 3D points online from 4-6 frames. Therefore, the primary drawback of the COLMAP baseline is that COLMAP requires more than \textbf{30 hours} to complete on a 128-thread CPU and 4$\times$ RTX A6000 GPUs for a single scene in TartanAir, a significantly longer time than our approach used. 

Without changing any of the evaluation protocols, 
we show the feature matching experiment result in \tref{tab:supp:colmap:tartanair-mma} and pose estimation experiment result in \tref{tab:supp:eth3d}. We find that COLMAP is insufficient to guarantee feature matching accuracy. This could be attributed to the MVS failing to provide accurate per-pixel depth value. However, COLMAP supervision is slightly more accurate on the pose estimation task, except that it fails on the \texttt{table} scene. This is because using more frames in the reconstruction would lead to a stronger geometric constraint and more accurate estimation of camera poses in case the scene is small. 

\begin{table*}[h]
  \small
  \setlength{\tabcolsep}{0.14em}
  \renewcommand{\arraystretch}{1.2}
        \caption{Comparison with COLMAP supervised baseline on the {feature matching task} } \label{tab:supp:colmap:tartanair-mma}
  \resizebox{\textwidth}{!}{
      \begin{tabular}
        {c | ccc !{\color{lightgray}\vrule} ccc | ccc!{\color{lightgray}\vrule}ccc!{\color{lightgray}\vrule}ccc|ccc!{\color{lightgray}\vrule}ccc|ccc
      }
      \toprule
      Scene & \multicolumn{3}{c}{Indoor} & \multicolumn{3}{c}{Outdoor} & \multicolumn{3}{c}{Natural} & \multicolumn{3}{c}{Artificial} & \multicolumn{3}{c}{Mixed} & \multicolumn{3}{c}{Easy} & \multicolumn{3}{c}{Hard} & \multicolumn{3}{c}{Overall} \\ 
      \midrule
      reproj thres & 1px & 2px & 5px & 1px & 2px & 5px & 1px & 2px & 5px & 1px & 2px & 5px & 1px & 2px & 5px & 1px & 2px & 5px & 1px & 2px & 5px & 1px & 2px & 5px \\
      \midrule
      COLMAP & 58.9 & 75.5 & 91.1 & 61.2 & 77.1 & 91.6 & 54.1 & 71.7 & 89.2 & 61.6 & 77.6 & 92.0 & 63.0 & 78.2 & 92.0 & 64.0 & 80.2 & 93.3 & 57.3 & 73.2 & 89.7 & 60.5 & 76.6 & 91.5 \\
      {iDKM (Ours)} & {62.9} & {78.9} & {93.1} & {66.6} & {81.4} & {94.4} & {60.7} & {77.5} & {92.6} & {65.9} & {80.9} & {94.1} & {68.0} & {82.6} & {94.9} & {69.6} & {84.5} & {95.8} & {60.6} & {76.4} & {92.1} & {65.5} & {80.7} & {94.0} \\
      \bottomrule
      \end{tabular}
      }
      \vspace{-10pt}
\end{table*}

\section{Evaluation on Larger Frame Rate}
The TartanAir dataset already includes aggressive camera motions. 
However, if frame spacing is enlarged to 8x, iMatching can still work, as 30\% of improvement on sequences from TartanAir is observed in \tref{tab:supp:tta:8x}.

\begin{table}[h]
    \small
    \centering
    \caption{Feature matching on TartanAir with larger frame rate}
    \begin{tabular}{c|ccc|ccc}
        \hline
        Method & \multicolumn{3}{c}{CAPS} & \multicolumn{3}{c}{iCAPS} \\
         & \multicolumn{3}{c}{(pretrained)} & \multicolumn{3}{c}{(finetuned)} \\
        \hline
        thres & 1px & 2px & 5px & 1px & 2px & 5px \\ 
        \hline
        Amusement Easy & 9.2 & 26.4 & 55.0 & 11.0 & 28.8 & 57.5\\
        Hospital Hard & 8.9 & 23.4 & 42.4 & 12.5 & 26.9 & 45.9\\
        \hline
    \end{tabular}
    \label{tab:supp:tta:8x}
\end{table}

\section{Limitations} Our proposed iMatching optimizes feature matching network parameters, camera pose, and 3D landmark locations jointly on contiguous video segments. Extending the work so that it is adaptable to arbitrary unordered image collection is an intriguing direction for future research. Furthermore, the research focuses on benchmarks specifically related to feature correspondence models. Investigating the effect of the training scheme on down-stream tasks such as localization, SfM, and SLAM would be an interesting future direction. 

\end{document}